\documentclass{article}
\usepackage{ijcai22}
\usepackage{float}
\usepackage{times}
\usepackage{soul}
\usepackage{url}
\usepackage{bm}
\usepackage[hidelinks]{hyperref}
\usepackage[utf8]{inputenc}
\usepackage[small]{caption}
\usepackage{graphicx}
\usepackage{amsmath}
\usepackage{amsthm}
\usepackage{booktabs}
\usepackage{algorithm}
\usepackage{algorithmic}
\usepackage{amssymb}
\usepackage{multirow}
\usepackage{enumitem}
\usepackage[switch]{lineno}
\title{Boosting Multi-Label Image Classification with Complementary Parallel Self-Distillation}

\author{
Jiazhi Xu$^1$\and
Sheng Huang$^1$\footnote{Corresponding Author}\and
Fengtao Zhou$^1$\and
Luwen Huangfu$^2$\and
Daniel Zeng$^3$\And
Bo Liu$^4$
\\
\affiliations
$^1$School of Big Data and Software Engineering, Chongqing University\\
$^2$Fowler College of Business \& Center for Human Dynamics in the Mobile Age, San Diego State University,
$^3$Institute of Automation, Chinese Academy of Sciences,
$^4$JD.com\\
\emails
\{xujiazhi, huangsheng, zft\}@cqu.edu.cn,
lhuangfu@sdsu.edu,
dajun.zeng@ia.ac.cn,
kfliubo@gmail.com
}

\begin{document}

\maketitle

\begin{abstract}
\vspace{-0.2cm}
Multi-Label Image Classification (MLIC) approaches usually exploit label correlations to achieve good performance.
However, emphasizing correlation like co-occurrence may overlook discriminative features of the target itself and lead to model overfitting, thus undermining the performance.
In this study, we propose a generic framework named Parallel Self-Distillation (PSD) for boosting MLIC models.
PSD decomposes the original MLIC task into several simpler MLIC sub-tasks via two elaborated complementary task decomposition strategies named Co-occurrence Graph Partition (CGP) and Dis-occurrence Graph Partition (DGP).
Then, the MLIC models of fewer categories are trained with these sub-tasks in parallel for respectively learning the joint patterns and the category-specific patterns of labels.
Finally, knowledge distillation is leveraged to learn a compact global ensemble of full categories with these learned patterns for reconciling the label correlation exploitation and model overfitting.
Extensive results on MS-COCO and NUS-WIDE datasets demonstrate that our framework can be easily plugged into many MLIC approaches and improve performances of recent state-of-the-art approaches.
The explainable visual study also further validates that our method is able to learn both the category-specific and co-occurring features. The source code is released at https://github.com/Robbie-Xu/CPSD.
\end{abstract}

\vspace{-0.45cm}
\section{Introduction}
\vspace{-0.05cm}
Natural images often contain multiple visual objects, which can be characterized by a set of image labels.
Multi-label image classification (MLIC) task is to recognize all these objects, which is highly relevant to other vision tasks such as object detection, image retrieval, and semantic segmentation.

Most existing MLIC research works focus on exploiting the label correlation property, which distinguishes it from the single-label image classification problem.
Label correlation exploitation strategies, such as pair-wise and high-order label correlation have been extensively studied.
Deep learning-based approaches, such as RNN~\cite{wang2016cnn}, graph model~\cite{chen2019learning,chen2019multi,nguyen2021modular} and attention mechanism~\cite{gao2021learning} are widely employed to encode the image label correlation, yielding decent performance.
\begin{figure}[t]
    \centering
    \vspace{-0.1cm}
    \includegraphics[scale=0.13]{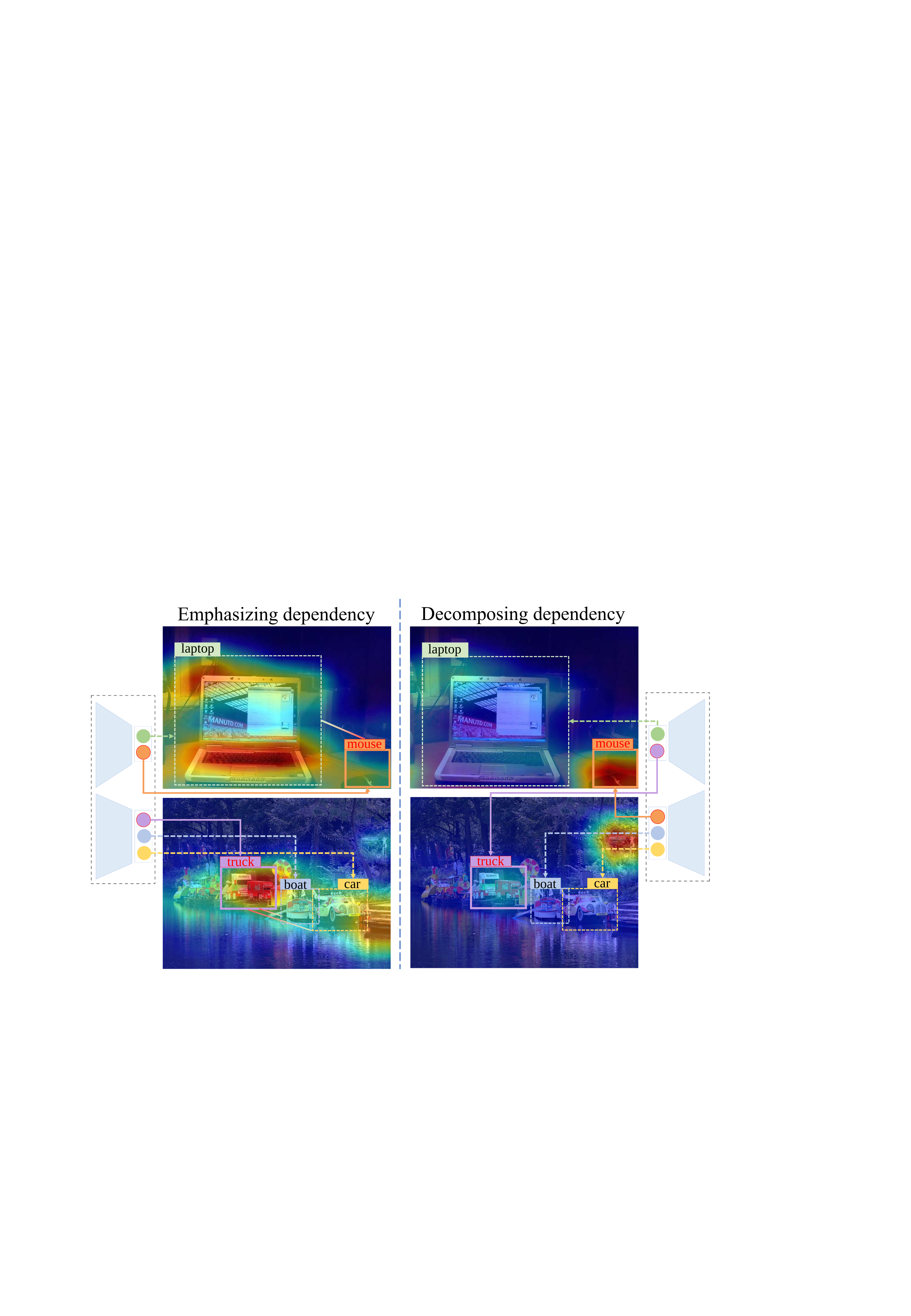}
    \vspace{-0.3cm}
    \caption{\textbf{Class Activation Map (CAM) of class \emph{mouse} and \emph{truck}}.
    Both target labels are marked in red.
    Overemphasizing the correlation by specifically putting the \emph{mouse} and \emph{laptop} categories together to train an MLIC model leads to model overfitting because the MLIC model infers the label \emph{mouse} only based on the features of a \emph{laptop}, e.g. upper left.
    On the contrary, if we train individual models parallelly for those two labels, the model will capture the category-specific features, e.g., upper right.
    However, when the object is not well exposed, the model needs to rely on the label-dependence knowledge to mine the category-specific feature.
    Training together succeeds in recognizing label \emph{truck} with the help of co-occurring \emph{car} and \emph{boat}, e.g., lower left.
    It is quite challenging to reconcile the model overfitting and the label correlation exploitation.
    }
    \label{fig1}
    \vspace{-0.5cm}
\end{figure}

Although label correlation is a useful feature for the MLIC problem, it can also be misleading due to model overfitting.
As illustrated by Figure 1, emphasizing co-occurrence leads to an inference of targets primarily from co-occurring objects, which may not represent all cases.
On the contrary, decomposing this correlation leads to learning discriminative features of the target itself, which may also fail in classification due to the lack of context.
Therefore, in addition to capturing label co-occurrence, our proposed framework also includes the discriminative features of individual labels, represented by dis-occurrence.
Furthermore, because of its multi-class nature, the problem complexity of an MLIC task, in terms of the prediction space grows exponentially as the category number increases, e.g., the prediction result has $2^c$ possibilities for a $c$-class MLIC problem.
When the image category number is higher, the model learning task becomes more challenging and issues with model overfitting are more likely to occur.
One prominent algorithm branch to reduce the complexity is to decompose the original problem into a set of binary problems with common strategies like one-vs-one and one-vs-all, or advanced strategies like D-Chooser~\cite{chen2021efficient}.

In this paper, we adopt this divide-and-conquer thought and propose a generic MLIC framework that addresses both model overfitting and label dependency modelling.
We first decompose an MLIC task into several simpler sub-tasks, each with fewer object categories.
Individual models are trained in parallel for tackling these sub-tasks.
After that, knowledge distillation, an effective way to learn a compact model with generalization capability from model ensemble~\cite{gou2021knowledge}, is conducted to learn a global model containing all object categories.
The label decomposition reduces the complexity of each sub-task, which helps each individual model learn more representative features.
In the model distillation process, those sub-models serve as teachers whose logit outputs are utilized as the soft targets to supervise the learning of a model which contains all categories with the same architecture as those teachers (i.e., self-distillation).
These soft targets function as a label smoothing regularizer~\cite{yuan2020revisiting} for a better optimization.
The contributions of our method are summarized as follows:

$\bullet$ A generic MLIC framework, called Parallel Self-Distillation (PSD) is proposed. With proper task-decomposition strategy, the original MLIC model training task is formulated as a multiple sub-model parallelly training task, then these sub-models are distilled into a global model.We demonstrate that our framework can be flexibly applied to existing MLIC models and improve their performances.

$\bullet$ Two strategies, namely Co-occurrence Graph Partition (CGP) and Dis-occurrence Graph Partition (DGP), are elaborated, for decomposing the MLIC task via label partition. They model the label correlation by two complementary graphs. The co-occurrence graph models the label correlation, based on which the spectral clustering result tends to assign co-occurring labels into the same cluster. This induces the individually trained sub-model to learn the joint pattern of these co-occurring classes. While the dis-occurrence graph assigns labels without co-occurrence into one task for learning the category-specific patterns. These two complementary strategies are simultaneously leveraged in PSD to reconcile the model overfitting and label correlation exploitation.

We conduct extensive experiments on two widely-used MLIC datasets, MS-COCO and NUS-WIDE.
Experimental results demonstrate that our framework can be plugged into different approaches to boost the performance without increasing complexity.
We also visualize the implicit attention of our framework to expose the overfitting of co-occurrence and demonstrate the effectiveness of our approach.

\vspace{-0.25cm}
\section{Related works}
\vspace{-0.05cm}
\paragraph{Multi-label Image Classification (MLIC)} is different from single-label image classification in that it relies on the label correlation property.
Many MLIC approaches have been devoted to exploiting this property.
For example, the methods proposed in~\cite{chen2019learning,chen2019multi} build label correlation graphs and adopt GNN for label feature learning.
Sample imbalance is another issue of MLIC.
In~\cite{wu2020distribution}, a distribution-balanced loss is proposed.
It re-balances the training sample weights and designs a negative-tolerant regularization which can avoid over-suppression caused by the dominance of negative classes.
An asymmetric loss is proposed in~\cite{ridnik2021asymmetric}, where the contribution of positive samples is maintained.
Vision Transformer~\cite{dosovitskiy2020image} has recently been introduced to MLIC not only because of its strong feature extraction capability, but also because the self-attention mechanism can capture rich patterns between visual features and class label tokens~\cite{lanchantin2021general,liu2021query2label}.
In this work, we test the performance of our proposed framework within Transformer with a naive Transformer encoder.

\paragraph{Knowledge Distillation} was initially proposed to transfer knowledge from large complex networks to slimmer networks in order to retain the performance of the large network with less computation and model size~\cite{hinton2015distilling}.
~\cite{zhang2019your} find that distilling a pretrained model with the same architecture can boost the model performance.
The technique is called self-distillation.
Zhou \emph{et al.} investigate the bias-variance tradeoff brought by distillation and propose to use weighted soft labels that enable a sample-wise bias-variance tradeoff~\cite{zhou2021rethinking}.
In~\cite{xiang2020learning}, a multiple experts distillation method is proposed to handle the long-tailed distribution in the image classification task.
The application of KD in MLIC can be found in~\cite{song2021handling}. In that work, model distillation is adopted to alleviate the model bias toward difficult categories.

\begin{figure*}[t]
    \centering
    \vspace{-0.2cm}
    \resizebox{!}{85mm}{
    \includegraphics[width=3\columnwidth]{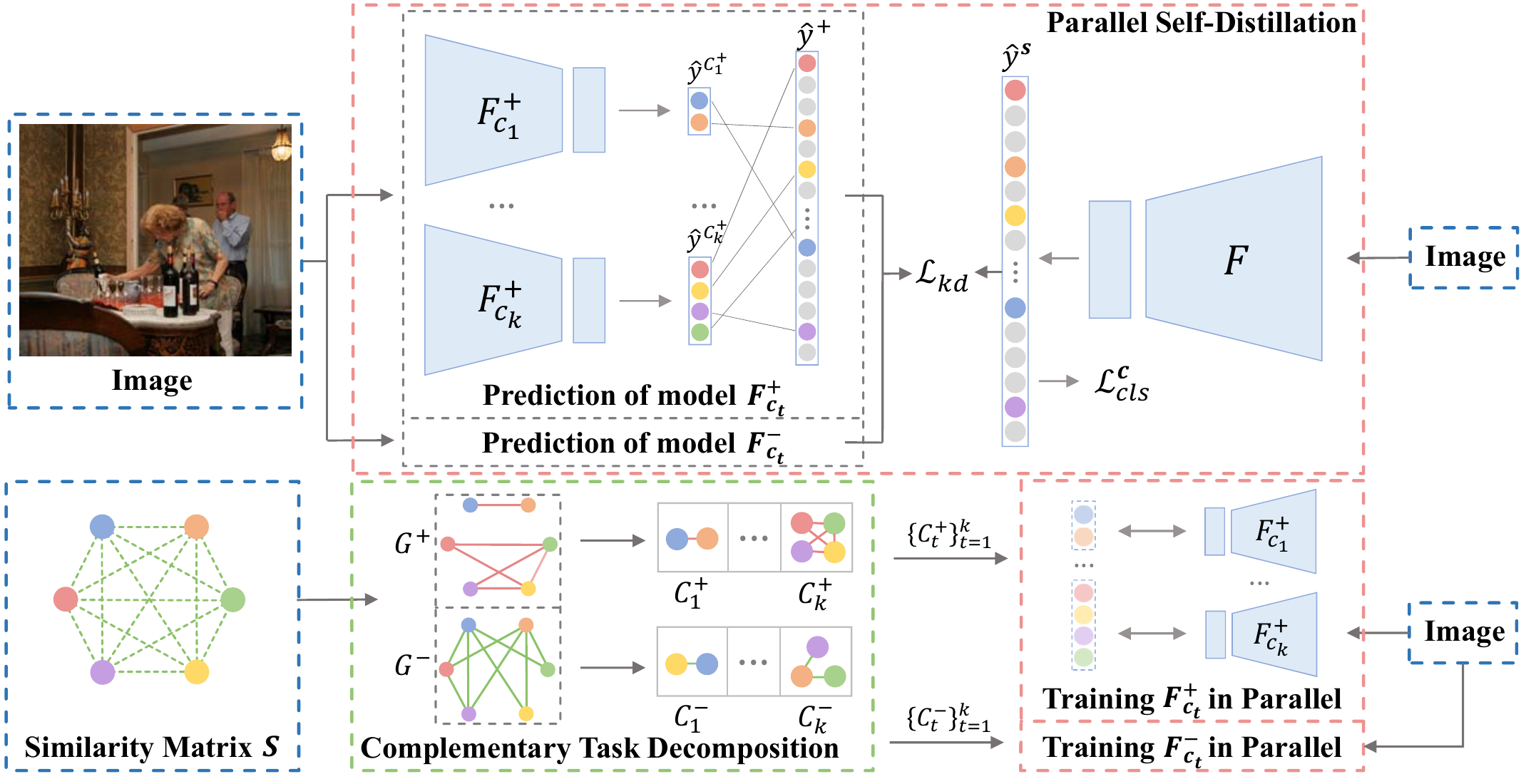}
    }
    \vspace{-0.3cm}
    \caption{\textbf{The overview of the PSD framework.}
    Blue, green, and red boxes indicate input, task decomposition and the PSD main flow respectively.
    The superscripts \emph{$+$} and \emph{$-$} respectively represent co-occurrence and dis-occurrence branches.
    The omitted operations in the dis-occurrence branch are the same as the ones in co-occurrence branch. Their only difference is their applied sub-tasks.
    }
    \label{fig2}
    \vspace{-0.5cm}
\end{figure*}

\vspace{-0.2cm}
\section{Methodology}
\vspace{-0.1cm}
\subsection{Preliminary and Overview}
Given a MLIC Task $\mathcal{T}:=\{(X,Y)\}$ where $X$ is the image set and $Y$ is its corresponding label set, the goal is to establish a visual learning model $F(\cdot)$, which is able to predict the labels of a given image $x \in X$, $y\leftarrow F(x)$. $y=[y(1),y(2),\cdots,y(m)]\in Y$ is a $m$-dimensional binary label vector and $m$ is the number of categories.
The binary element $y(j)\in \{0,1\}$ indicates the existence of the corresponding category in an image.
For example, if $y_{i}(j)=1$, the sample $x_i$ contains the $j$-th category, otherwise not.
Let $\mathcal{C}=\{c_1,c_2,\cdots, c_m\}$ be the category set where $|\mathcal{C}|=m$.
With regard to the multi-class classification issue, a larger $m$ implies a high-dimensional label space and thus a more difficult MLIC task.
MLIC often suffers from more complexity in comparison to ordinary single-label multi-class image classification even in the same label space because of the label correlation.
The divide-and-conquer strategy is an intuitive and common way for tackling such complex tasks.
The basic idea of this strategy is to decompose the complex task into a set of simpler sub-tasks, and then assemble the sub-solutions to yield the final solution of the original task.

In this paper, we follow such a strategy and propose a Parallel Self-Distillation (PSD) framework for addressing MLIC issue. The architecture of PSD is shown in Figure~\ref{fig2}.
In PSD, the first step is to decompose the original MLIC task $\mathcal{T}$ into several simple MLIC sub-tasks by dividing the category set into several smaller subsets with a decomposition strategy  $\{T_t\}_{t=1}^k=\Psi(\mathcal{T},\mathcal{C})$ where $k$ is the number of sub-tasks and $\Psi(\cdot,\cdot)$ is a task decomposition strategy.
$T_t:=\{(X_{C_t}, Y_{C_t})|C_t \in \mathcal{C}\}$ is the $t$-th sub-task where $C_t$ is a subset of $\mathcal{C}$ and is the label space of $T_t$. $X_{C_t}$ is the images, which contain categories in $C_t$, and $Y_{C_t}$ is the label set of these corresponding images.
The second step is to train an MLIC model $F_{C_t}(\cdot)$ for each sub-task individually.
Finally, these trained MLIC models are assembled to yield a final MLIC model which considers them as teachers for knowledge distillation.
Moreover, we elaborate two task decomposition strategies named Co-occurrence Graph Partition (CGP) and Dis-occurrence Graph Partition (DGP) for boosting PSD.
The decomposition issue is regarded as a spectral clustering problem.
In CGP, a label co-occurrence graph is constructed and clustered to assign training samples containing the co-occurring labels into the same sub-task to induce the model to learn the joint patterns.
On the contrary, DGP constructs a label dis-occurrence graph for spectral clustering, and tends to assign labels without co-occurrence into the same sub-task to better learn the category-specific patterns.
These two strategies are intrinsically complementary with each other and will be applied together to PSD with respect to both aspects of the information, namely Complementary PSD (CPSD).
It is worth noting that this proposed approach can be flexibly plugged into any deep learning-based MLIC models for further improvement.

\vspace{-0.2cm}
\subsection{Complementary Task Decomposition}
\vspace{-0.05cm}
Task decomposition is the key of the PSD framework.
We intend to reduce the dimension of label space to simplify the MLIC task by considering the task decomposition problem as a label partition (or category clustering) issue.
In the design of the MLIC task composition strategy, two important aspects should be paid attention to.
One is the simplification of the task complexity.
The other is the label correlation exploitation.
In model optimization, these two aspects are somewhat conflicted.
The label correlation complicates the MLIC task and easily triggers model overfitting to mislead the feature learning to learn features of the co-occurring object instead of the original ones.

In order to avoid this, we design two task decomposition strategies named Co-occurrence Graph Partition (CGP) and Dis-occurrence Graph Partition (DGP) based on spectral clustering.
Finally, these two aspects of knowledge will be distilled into a unified MLIC model.
CGP and DGP translate the task decomposition issue as the unsupervised category graph partition problem for solution.
We construct a \emph{co-occurrence graph} $G^+$ and a \emph{dis-occurrence graph} $G^-$ to encode the joint and specific patterns among categories respectively.
By considering categories $\mathcal{C}$ as vertices and measuring the co-occurrence probabilities of categories as their similarities, we can define a $m\times m$-dimensional similarity matrix $S$.
Its $ij$-th element is $S_{ij}=e_{ij}/n_i \in [0,1]$ where $e_{ij}$ is the amount of images containing both $c_i$ and $c_j$. And $n_i$ is the amount of images containing $c_i$.
Mathematically speaking, let $P$ be the affinity matrix of category graph $G$. The affinity matrices of $G^+$ and $G^-$ can be denoted as follows,
{\small
\begin{equation}\label{smooth_sym}
\setlength{\abovedisplayskip}{1mm}
\setlength{\belowdisplayskip}{1mm}
P=\left\{
\begin{aligned}
  P^+&=\frac{(\sqrt[\tau]{S}+\sqrt[\tau]{S}^T)}{2}, & G=G^+ \\
  P^-&=I-\frac{(\sqrt[\tau]{S}+\sqrt[\tau]{S}^T)}{2}, & G=G^-
\end{aligned}
\right.
\end{equation}
}
where $\tau$ is a positive hyper-parameter for smoothing the co-occurrence probabilities which follow the long-tail distribution. A higher value of  $\tau$ is able to alleviate more suppression of the head to the tail.  Since $G^+$ and $G^-$ are the undirected graphs, we also symmetrize the affinity matrices. The affinity matrices $P^+$ and $P^-$ respectively encode the degrees of co-occurrence and dis-occurrence among categories. Therefore, we can use them to produce graph Laplacians, and then conduct spectral clustering to partition the label space as a normalized graph cut problem,
{\small
\begin{equation}
\setlength{\abovedisplayskip}{1mm}
\setlength{\belowdisplayskip}{1mm}
    \hat{F}\leftarrow\mathop{\arg\min}\limits_{F} \text{Trace}(F^{T}D^{-\frac{1}{2}}LD^{-\frac{1}{2}}F),~\text{s.t.}~F^{T}F=I,
\end{equation}
}
where $L=D-P$ is the Laplacian matrix and $D$ is the degree matrix of $G$. $D$ is a diagonal matrix whose $ii$-th element $D_{ii}=\sum_j P_{ij}$. $F$ is the learned graph embedding of vertices (categories), which encode the co-occurrence or dis-occurrence information of each category. Each of its columns encodes a graph cut operation represented by a vector. The aforementioned optimization problem can be efficiently solved by eigenvalue decomposition. The optimal $\hat{F}$ is the eigenvectors corresponding to the top-$k$ minimum eigenvalues. By employing $k$-means to $\hat{F}$, the categories can be clustered into $k$ category subsets $\{C_t\}_{t=1}^k$ where $\bigcup_{t=1}^kC_t=\mathcal{C}$ and $~\bigcap_{t=1}^k C_t= \oslash$. Then the task can be decomposed into $k$ sub-tasks via data sampling according to the clustered category subset, $\{T_t\}_{t=1}^k$.
The sub-tasks generated by CGP $\{T^+_t\}_{t=1}^k$ are used to induce the model to learn the features of co-occurring objects. The sub-tasks generated by DGP $\{T^-_t\}_{t=1}^k$ exhibit lower complexity because of the neglect of the label correlation and encourage the model to focus on the extraction of category-specific features. All the teacher models trained by these sub-tasks will finally be used in the Parallel Self-Distillation (PSD), $y\underset{(x,y)\in T_t}\leftarrow \sigma(F_{C_t}(x))$ where $\sigma$ means sigmoid activation, $T_t \in \{T^-_t\}_{t=1}^k\cup\{T^+_t\}_{t=1}^k$ since we expect to reconcile both aspects of knowledge and exploit their complementarity natures for better supervising the training of the student model. Moreover, the teacher models trained by these sub-tasks and generated from both strategies are highly different, which can further benefit PSD from the perspective of ensemble learning.

\vspace{-0.15cm}
\subsection{Parallel Self-Distillation}
\vspace{-0.05cm}

For each sub-task $T_t$, we train a teacher model,
\begin{equation}\label{teacher_train}
\setlength{\abovedisplayskip}{3pt}
\setlength{\belowdisplayskip}{1mm}
\hat{F}_{C_t}\leftarrow\arg\underset{F_{C_t}}\min~L^{C_t}_{cls}
\end{equation}
with the Asymmetric Loss (ASL)~\cite{ridnik2021asymmetric} as the classification loss to suppress the negative effects from superabundant negative samples,
{\small
\begin{equation}\label{asl}
    \setlength{\abovedisplayskip}{3pt}
    \setlength{\belowdisplayskip}{1mm}
        L^{C_t}_{cls}=\!\sum_{x_i \in X_{C_t}}\!\sum_{c_j\in C_t}\!
        \left\{
            \begin{aligned}
                \!(1-\hat{y}_i^{C_t}(j))^{\gamma_+}\!\log(\hat{y}_i^{C_t}(j)),&~y_i(j)\!=\!1\\
                \!\hat{\bm{y}}_i^{C_t}(j)^{\gamma_-}\log(1-\hat{\bm{y}}_i^{C_t}(j)),&~y_i(j)\!=\!0\\
            \end{aligned}
        \right.
    \end{equation}}
where $y_i$ the ground truth label of $x_i$ and $\hat{y}_i^{C_t}=\sigma(F_{C_t}(x_i))$ is its predictions with respect to the category subset $C_t$. Its $j$-th element $\hat{y}_i^{C_t}(j)\in [0,1]$ is the predicted label (probability) of the sample $x_i$ with respect to the category $c_j$, and ${\bm{\hat{y}}}_i^{C_t}(j)=\max(\hat{y}_i^{C_t}(j)-\mu, 0)$. $\mu\geq 0$ is a threshold to filter out the negative samples with low predictions. $\gamma_+\geq 0$ and $\gamma_-\geq 0$ are respectively the positive and negative focusing hyper-parameters defined in ASL.

Once all teacher models have been obtained, we can merge their prediction results to yield a complete logit prediction according to the category order,
\begin{equation}\label{}
\setlength{\abovedisplayskip}{1mm}
\setlength{\belowdisplayskip}{1mm}
\hat{y}_i=\rho(\{\hat{F}_{C_t}(x_i)|x_i\in X_{C_t}\}_{t=1}^k),
\end{equation}
where $\rho(\cdot)$ is a label merging and reshuffling operation. Here, let $\hat{y}_i^+$ and $\hat{y}_i^-$ be the logit predictions produced by the teacher models based on CGP and DGP respectively.
We establish a student model $F(\cdot)$ with the same architecture as teachers to produce the logit predictions of a sample, $\hat{y_i}^{\bm{s}}=F(x_i)$.
The Mean Square Error (MSE) is adopted to measure the discrepancy between the logit predictions of the student model and the teacher models,
\begin{equation}
\setlength{\abovedisplayskip}{1mm}
\setlength{\belowdisplayskip}{1mm}
    L_{kd} = \frac{1}{2}\sum_{i}\{||\hat{y}_i^{\bm{s}}-\hat{y}_i^+||_2^2+||\hat{y}_i^{\bm{s}}-\hat{y}_i^-||^2_2\},
\end{equation}
as the knowledge distillation loss for supervising the student model training.

Finally, the optimal student model can be obtained by addressing the following optimization problem,
\begin{equation}\label{student_train}
\setlength{\abovedisplayskip}{3pt}
\setlength{\belowdisplayskip}{1mm}
\hat{F}\leftarrow\arg\underset{F}\min~L^{\mathcal{C}}_{cls}+L_{kd},
\end{equation}
where $L^{\mathcal{C}}_{cls}$ is the ASL loss of the original task $\mathcal{T}$. $L^{\mathcal{C}}_{cls}$ can be constructed with Equation~\ref{asl} based on the full data $X$.
By the above manner, the learned optimal MLIC model $\hat{F}$ will finally incorporate the knowledge acquired by the CGP and DGP-based teacher models, and then we can use it to infer the label of a multi-label image in the testing stage, $y=\sigma(\hat{F}(x))$.


\begin{table*}[t]
    \centering
    \renewcommand{\arraystretch}{0.7}
    \resizebox{!}{55mm}{
    \begin{tabular}{lccccccccc}
        \toprule
        Methods                                  & Backbone       & Resolution      & mAP          & CP           & CR           & CF1          & OP           & OR           & OF1   \\
        \midrule
        ResNet-101~\cite{he2016deep}             & ResNet101      & 224$\times$224  & 78.3         & 80.2         & 66.7         & 72.8         & 83.9         & 70.8         & 76.8  \\
        DSDL~\cite{zhou2021deep}                 & ResNet101      & 448$\times$448  & 81.7         & 84.1         & 70.4         & 76.7         & 85.1         & 73.9         & 79.1  \\
        CPCL~\cite{zhou2021multi}                & ResNet101      & 448$\times$448  & 82.8         & 85.6         & 71.1         & 77.6         & 86.1         & 74.6         & 79.9  \\
        ML-GCN~\cite{chen2019multi}              & ResNet101      & 448$\times$448  & 83.0         & 85.1         & 72.0         & 78.0         & 85.8         & 75.4         & 80.3  \\
        KSSNet~\cite{liu2018multi}               & ResNet101      & 448$\times$448  & 83.7         & 84.6         & 73.2         & 77.2         & 87.8         & 76.2         & 81.5  \\
        MS-CMA~\cite{you2020cross}               & ResNet101      & 448$\times$448  & 83.8         & 82.9         & 74.4         & 78.4         & 84.4         & 77.9         & 81.0  \\
        MCAR~\cite{gao2020multi}                 & ResNet101      & 448$\times$448  & 83.8         & 85.0         & 72.1         & 78.0         & 88.0         & 73.9         & 80.3  \\
        Q2L-R101~\cite{liu2021query2label}       & ResNet101      & 448$\times$448  & 84.9         & 84.8         & 74.5         & 79.3         & 86.6         & 76.9         & 81.5  \\
        \midrule
        ResNet101{$^*$}(baseline)                & ResNet101      & 448$\times$448  & 81.6         & 80.6         & 72.7         & 76.4         & 83.7         & 76.7         & 80.0\\
        \textbf{Ours\,+\,ResNet101}              & ResNet101      & 448$\times$448  & 83.1         & 83.5         & 73.6         & 78.2         & 84.8         & 77.3         & 80.9\\
        \midrule
        ResNet101\,+\,TF\textbf{$^*$}            & ResNet101      & 448$\times$448  & 84.3         & 87.4         & 71.6         & 78.7         & 87.9         & 75.2         & 81.0  \\
        \textbf{Ours\,+\,ResNet101\,+\,TF}       & ResNet101      & 448$\times$448  & \textbf{85.2}& 84.9         & 75.5         & \textbf{79.9}& 85.6         & 78.5         & \textbf{81.9}\\
        \midrule
        Q2L-R101\textbf{$^*$}                    & ResNet101      & 448$\times$448  & 84.0         & 82.0         & \textbf{75.8}& 78.8         & 83.3         & \textbf{78.8}& 81.0   \\
        \textbf{Ours\,+\,Q2L-R101}               & ResNet101      & 448$\times$448  & 84.9         & \textbf{88.4}& 71.7         & 79.2         & \textbf{89.3}& 74.8         & 81.4  \\
        \bottomrule
        \toprule
        SSGRL~\cite{chen2019learning}            & ResNet101      & 576$\times$576  & 83.8         & \textbf{89.9}& 68.5         & 76.8         & \textbf{91.3}& 70.8         & 79.7  \\
        C-Trans~\cite{lanchantin2021general} & ResNet101      & 576$\times$576  & 85.1         & 86.3         & 74.3         & 79.9         & 87.7         & 76.5         & 81.7  \\
        ADD-GCN~\cite{ye2020attention}           & ResNet101      & 576$\times$576  & 85.2         & 84.7         & 75.9         & 80.1         & 84.9         & 79.4         & 82.0  \\
        Q2L-R101~\cite{liu2021query2label}       & ResNet101      & 576$\times$576  & 86.5         & 85.8         & 76.7         & 81.0         & 87.0         & 78.9         & 82.8  \\
        \midrule
        ResNet101\,+\,TF\textbf{$^*$}            & ResNet101      & 576$\times$576  & 85.9         & 88.6         & 73.4         & 80.3         & 88.8         & 76.8         & 82.4  \\
        \textbf{Ours\,+\,ResNet101\,+\,TF}       & ResNet101      & 576$\times$576  & \textbf{86.7}& 83.5         & \textbf{79.0}& \textbf{81.2}& 84.5         & \textbf{81.4}& \textbf{82.9}\\
        \bottomrule
        \toprule
        TResL~\cite{ridnik2021asymmetric}        & TResNetL       & 448$\times$448  & 86.6         & 87.2         & 76.4         & 81.4         & 88.2         & 79.2         & 81.8  \\
        Q2L-TResL~\cite{liu2021query2label}      & TResNetL       & 448$\times$448  & \textbf{87.3}& \textbf{87.6}& 76.5         & 81.6         & \textbf{88.4}& 79.2         & 81.8  \\
        \midrule
        TResL\textbf{$^*$}(baseline)             & TResNetL       & 448$\times$448  & 86.2         & 85.0         & 77.5         & 81.1         & 85.6         & 80.4         & 82.9  \\
        \textbf{Ours\,+\,TResL}                  & TResNetL       & 448$\times$448  & \textbf{87.3}& 85.5         & \textbf{78.9}& \textbf{82.1}& 85.7         & \textbf{81.5}& \textbf{83.7}\\
        \bottomrule
        \toprule
        ML-GCN~\cite{nguyen2021modular}          & ResNeXt50-SWSL & 448$\times$448  & 86.2         & 85.8         & 77.3         & 81.3         & 86.2         & 79.7         & 82.8      \\
        MGTN~\cite{nguyen2021modular}            & ResNeXt50-SWSL & 448$\times$448  & 87.0         & 86.1         & 77.9         & 81.8         & 87.7         & 79.4         & 83.4      \\
        \midrule
        ResNeXt50\textbf{$^*$}(baseline)         & ResNeXt50-SWSL & 448$\times$448  & 86.7         & 85.8         & 77.8         & 81.6         & 86.9         & 80.3         & 83.5          \\
        \textbf{Ours\,+\,ResNeXt50}              & ResNeXt50-SWSL & 448$\times$448  & \textbf{87.7}& \textbf{86.9}& \textbf{78.6}& \textbf{82.5}& \textbf{87.6}& \textbf{80.9}& \textbf{84.1} \\
        \bottomrule
    \end{tabular}
    }
    \vspace{-0.2cm}
    \caption{The MLIC performances of different methods on MS-COCO datasets with pretrained backbones on ImageNet-1k.
   \emph{*} indicates the results reproduced by the corresponding released codes or their modified versions. The best results for each backbone are in \textbf{bold.}
    }
    \label{coco}
    \vspace{-0.5cm}
\end{table*}

\vspace{-0.25cm}
\section{Experiments}
\vspace{-0.1cm}
\subsection{Experimental Setup}
\paragraph{Datasets.}
Two widely used MLIC datasets, named MS-COCO and NUS-WIDE, are used for the evaluation of our method.
MS-COCO contains 122,218 images with 80 categories of objects in natural scenes, including 82,081 images for training and 40,137 images for validation.
In the official partition of NUS-WIDE dataset, it contains 125,449 labeled training pictures and 83,898 labeled test pictures from Flickr, which share 81 labels in total.

Following the conventions, mAP (mean average precision) is deemed as the main evaluation metric. We also report overall precision (OP), recall (OR), F1-measure (OF1) and per-category precision (CP), recall (CR), F1-measure (CF1).

\paragraph{Implementation Details.}
We conduct experiments on three popular backbones, namely ResNet101~\cite{he2016deep}, TResNetL1~\cite{ridnik2021asymmetric} and ResNeXt50-SWSL~\cite{yalniz2019billion}, which are all pretrained on ImageNet-1K.
A naive Vision Transformer encoder ~\cite{dosovitskiy2020image} named ResNet101-TF is implemented with visual tokens extracted from ResNet101. In ResNet101-TF, $m$ class tokens are extracted from GloVe~\cite{pennington2014glove} for class predictions, where the depth, number of multi-heads attention, and hidden dimensions are set to be 3, 4, and 1024 respectively.
Q2L~\cite{liu2021query2label} is also adopted to verify the effect of our approach on well-designed methods.
All experiments follow a training pipeline where Adam optimizer is used with weight decay of $10^{-4}$ under a batch size of 32. ASL is applied as the default classification loss function, and the hyper-parameters of ASL are simply left as their default settings. $\tau$ in Equation~\ref{smooth_sym} is set to be 3.
We set the training epoch to be 20 and 80 for sub-models and the compact global model individually.
The number of clusters $k$ will be discussed in our ablation study.

\vspace{-0.1cm}
\subsection{Comparison with State-of-The-Art Methods}
Tables~\ref{coco} and~\ref{nus} report the multi-label image classification performances of several methods evaluated on MS-COCO and NUS-WIDE datasets respectively.
We use ResNet101 as the main baseline, while applying our proposed framework on recent benchmarks to evaluate the effectiveness.

The observations show that our method generally boosts all baselines, and performs the best on both datasets under different backbones and image resolutions. For example, the enhanced versions of ResNet101, ResNet101-TF, Q2L-R101, TResL and ResNeXt50 get 1.5\%, 0.9\%, 0.9\%, 1.1\% and 1.0\% gains respectively in mAP over their original ones on MS-COCO dataset. Such gains of ResNet101+TF and TResL are 1.7\% and 1.8\% on NUS-WIDE, which is a larger scale dataset. These experimental results also imply that our method performs much better on a larger-scale dataset.

In addition, the observations also imply that, based on our framework, the naive model is able to achieve state-of-the-art performances without involving additional costs in parameter scale or more complicated architectures. For example, our method gets 0.7\% gains in mAP over MGTN using only its backbone on MS-COCO dataset. Another interesting phenomenon is that we achieve smaller performance gains on the more advanced models. For example, CPSD improves more on ResNet101 in comparison with ResNet101-TF and Q2L-R101. We attribute this to the fact that the more powerful approaches are much harder to trap in model overfitting. Even so, our method still introduces a considerable improvement.


\begin{table}[t]
    \centering
    \resizebox*{!}{48mm}{
    \renewcommand{\arraystretch}{0.8}
        \begin{tabular}{lcccc}
            \toprule
            Methods                                 & mAP                  & CF1    & OF1   \\    
            \midrule
            MS-CMA~\cite{you2020cross}              & 61.4                 & 60.5   & 73.8  \\    
            SRN~\cite{zhu2017learning}              & 62.0                 & 58.5   & 73.4  \\    
            CPCL~\cite{zhou2021multi}               & 62.3                 & 59.2   & 73.0  \\    
            CADM~\cite{chen2019multi_}              & 62.8                 & 60.7   & 74.1  \\    
            Q2L-R101~\cite{liu2021query2label}      & 65.0                 & 63.1   & 75.0  \\    
            \midrule
            ResNet101+TF$^*$                        & 64.1                 & 62.8   & 74.9    \\    
            \textbf{Ours+ResNet101+TF}              & \textbf{65.8}        & \textbf{64.0}    & \textbf{75.3}     \\    
            \bottomrule
            \toprule
            TResL~\cite{ridnik2021asymmetric}       & 65.2                 & 63.6   & 75.0  \\    
            Q2L-TResL~\cite{liu2021query2label}     & 66.3                 & 64.0   & 75.0  \\    
            \midrule
            TResL$^*$(baseline)                     & 64.7                 & 63.7   & 75.0  \\    
            \textbf{Ours+TResL}                     & \textbf{66.5}        & \textbf{64.6} & \textbf{75.5} \\    
            \bottomrule
        \end{tabular}
    }
    \vspace{-0.2cm}
    \caption{The MLIC performances of different methods on NUS-WIDE datasets with pretrained backbones on ImageNet-1k where the image resolution is 448$\times$448.\emph{*} indicates the results reproduced by the corresponding released codes or their modified versions.
    The best results are in \textbf{bold.}}
    \label{nus}
    \vspace{-0.4cm}
\end{table}

\vspace{-0.1cm}
\subsection{Ablation Study}

\begin{figure}[t]
    \centering
    \includegraphics[width=1\columnwidth]{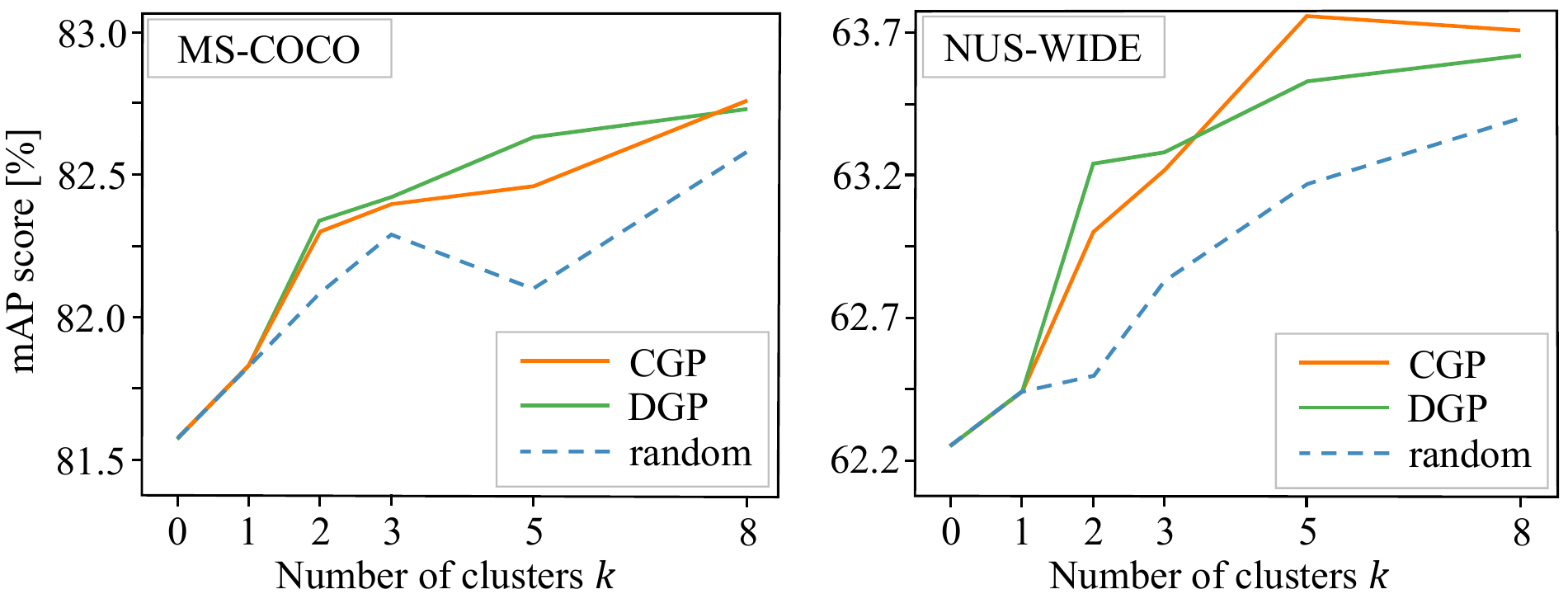}
    \vspace{-0.6cm}
    \caption{The performances of ResNet101 enhanced by PSD under different $k$ with different strategies on both MS-COCO and NUS-WIDE.
    CGPD, DGPD mean Co/Dis-Occurrence Graph Partition Distillation, which indicate student models that use only CGP or DGP decomposition strategy.
    0 and 1 respectively represent the baseline and the baseline with self-distillation.
    We conduct 3 independent experiments on random strategy and report the mean of their results for eliminating the effects of randomness.
    }
    \label{fig3}
    \vspace{-0.2cm}
\end{figure}

\paragraph{Discussion on Task Decomposition Strategy.}We plot the performances of PSD under different $k$ with different task decomposition strategies $\Psi$ on both MS-COCO and NUS-WIDE in Figure~\ref{fig3}.
The results indicate that the performances of PSD increased along with the increase of $k$ with all strategies on both datasets.
Moreover, our proposed strategies consistently perform much better than the random ones when $k\geq2$.
DGPD performs better than CGPD when $k$ is small while the best performances of the two strategies are highly similar.
We attribute this to the fact that overemphasizing the label correlation causes model overfitting more easily when the label space is in high-dimensional, while the label correlation is still able to benefit MLIC when the category-specific features are well learned.
Actually, reducing the size of the cluster (increasing $k$) can also be deemed as a natural way to break down label correlation.
We also notice that performance growth tends to slow down when $k\geq5$.
A larger $k$ means more teacher models are needed to be trained, which leads to higher time cost. To achieve the tradeoff between the performance and the model training time, we set $k = 5$.

\paragraph{Ablation Study on Components.} Table~\ref{ablation} shows the ablation study results of our method with different baselines on MS-COCO.
The results show that our PSD framework boosts all the baselines and outperforms the common Self-Distillation (SD) with considerable advantages using CGP, DGP or their combination.
For example, CPSD further improves SD by 1.4\%, 0.8\%, 0.8\%, 0.7\% and 0.8\% on ResNet101, TRestL, ResNeXt50, Q2L-R101 and ResNext101-TF respectively.
Moreover, CPSD performs much better than models using only one decomposition strategy, i.e. CGPD and DGPD.
The improvements under different baselines are around 0.5\%. These observations confirm the effectiveness of our method.

\begin{table}[t]
    \centering
    \resizebox{\columnwidth}{!}{
    \renewcommand{\arraystretch}{0.8}
        \begin{tabular}{lccccc}
            \toprule
                                    & R101      & TResL     & ResX50    & Q2L-R101  & R101+TF   \\
            \midrule
            baseline                & 81.6      & 86.2      & 86.7      & 84.0      & 84.3      \\
                \ + SD              & 81.9      & 86.5      & 86.9      & 84.2      & 84.4      \\
                \ + CGPD            & 82.4      & 86.6      & 87.3      & 84.6      & 84.7      \\
                \ + DGPD            & 82.7      & 86.8      & 87.4      & 84.3      & 84.5      \\
                \ + CPSD            & 83.1      & 87.3      & 87.7      & 84.9      & 85.2      \\
            \bottomrule
        \end{tabular}
    }
    \vspace{-0.2cm}
    \caption{\textbf{Component Ablation study of CPSD.}
    SD, CGPD, DGPD, CPSD mean Self-Distillation, Co/Dis-Occurrence Graph Partition Distillation and Complementary Parallel Self-Distillation respectively. Here, $k=5$.
    }
    \label{ablation}
    \vspace{-0.3cm}
\end{table}

\begin{figure}[t]
    \centering
    \includegraphics[width=1\columnwidth]{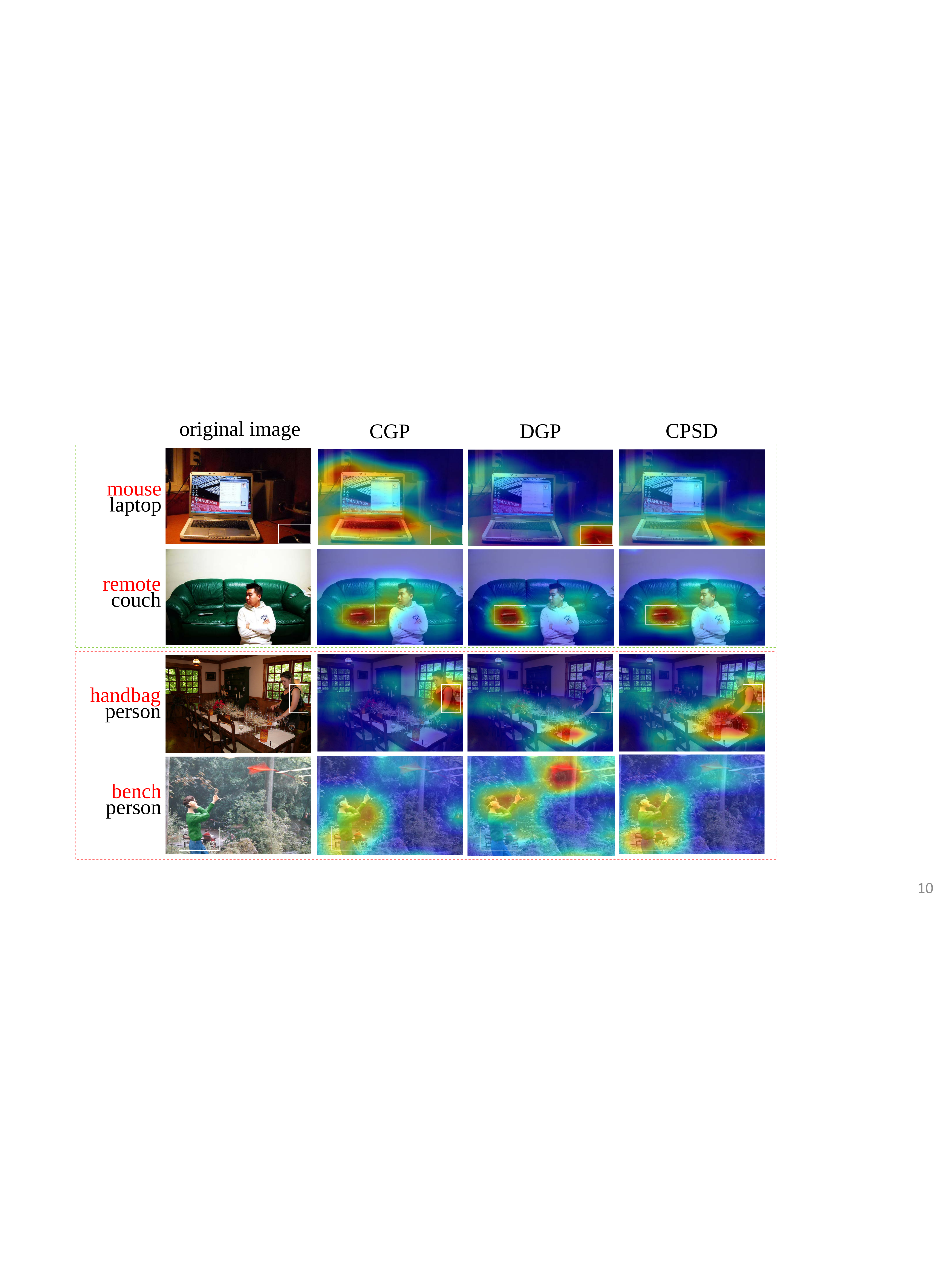}
    \vspace{-0.6cm}
    \caption{\textbf{Activation map visualizations of images under different decomposition strategies.}
    \emph{CGP, DGP} columns are the activation map visualizations of corresponding sub-model $F_{C_t}^+$ and $F_{C_t}^-$.
    \emph{CPSD} column is the visualizations of our final model.
    The red label indicates the target class we want to activate, while the others indicate the co-occurring classes.
    }
    \label{fig4}
    \vspace{-0.4cm}
\end{figure}

\subsection{Explainable Visualization Study}
We apply Class Activation Map (CAM) on the validation set of MS-COCO to visualize the implicit attentions of different images with respect to a specific category using different task decomposition strategies as shown in Figure~\ref{fig4}.
In this figure, the columns from left to right respectively show the original images and the images highlighted by the activations maps of the baseline ResNet101 models enhanced with PSD under CGP, DGP and their combination (CPSD).
The first two cases (the top two rows) show that the category-specific features have been ignored by the model based on CGP due to the model overfitting caused by the overemphasis of the label correlation, while they are well learned by our DGP models.
The last two cases show that the label co-occurrence information has not been exploited by the model based on DGP due to the occlusion, however the co-occurring categories are able to assist the model in discovering these features with the help of the label correlation information.
The visualizations demonstrate that our proposed method can perform well in each of these situations and also reflect that our method can better exploit and reconcile the category-specific knowledge and label correlation knowledge.

\vspace{-0.2cm}
\section{Conclusion}
\vspace{-0.05cm}
In this paper, we proposed a simple yet effective Parallel Self-Distillation (PSD) framework for Multi-Label Image Classification (MLIC). In this framework, the original complex MLIC task is decomposed into a set of simpler sub-tasks via label partition. Then multiple teacher models are trained in parallel to address these sub-tasks individually. The final model is obtained through the ensemble of these teacher models with knowledge distillation. For better boosting PSD, we introduce two task decomposition strategies, which address the task decomposition issue through conducting two complementary co-occurrence graph partitions. These two strategies, which respectively induce the models to learn the category-specific and category-correlated knowledge, are applied to PSD to set up the sub-tasks and together reconcile the two kinds of knowledge. Extensive experimental results on MS-COCO and NUS-WIDE demonstrate that our framework can be plugged into different approaches to boost the performances. The explainable visualization study also confirms the effectiveness of our method in feature learning.
\section*{Acknowledgements}
Reported research is partly supported by the National Natural Science Foundation of China under Grant 62176030 and 71621002,   the Natural Science Foundation of Chongqing under Grant cstc2021jcyj-msxmX0568, and the Strategic Priority Research Program of Chinese Academy of Sciences Grant XDA27030100.
\small
\bibliographystyle{named}
\bibliography{ijcai22}

\end{document}